\documentclass[10pt]{article} % 10pt font size is common

\usepackage[utf8]{inputenc}
\usepackage[T1]{fontenc}
\usepackage[margin=1in]{geometry} % Standard 1-inch margins
\usepackage{multicol}             % For creating two columns
\usepackage{helvet}               % Use Helvetica font (sans-serif)
\usepackage{graphicx}             % For images
\usepackage{amsmath}              % For math
\usepackage[dvipsnames]{xcolor}
\usepackage{caption}
\usepackage{tabularx}
\usepackage{booktabs} % For the nice lines
\usepackage{algorithm}
\usepackage{algorithmicx}
\usepackage{algpseudocode}
\usepackage{float}

\usepackage[numbers,sort&compress]{natbib} % For numbered, compressed citations [1]-[3]
\usepackage{booktabs}             % For professional-looking tables
\usepackage{hyperref}             % For clickable links and citations
\usepackage{float}    % For the [H] placement option to force figures here
\hypersetup{
    colorlinks=true,
    linkcolor=blue,
    filecolor=magenta,      
    urlcolor=cyan,
    citecolor=black,
}

\usepackage{authblk}
\title{\bfseries\huge OPTIC-ER: A Reinforcement Learning Framework for Real-Time Emergency Response and Equitable Resource Allocation in Underserved African Communities} % Bold and Huge title
\author{\large Mary Tonwe  } % Large font for author
\date{} % No date

\usepackage{titlesec}
\titleformat{\section}{\normalfont\scshape\bfseries}{\thesection.}{1em}{} % Small caps, bold for section titles
\titleformat{\subsection}{\normalfont\bfseries}{\thesubsection}{1em}{} % Bold for subsection titles
\titleformat{\subsubsection}{\normalfont\itshape}{\thesubsubsection}{1em}{}
\titlespacing*{\section}{0pt}{2.0ex plus 1ex minus .2ex}{1.0ex plus .2ex}
\titlespacing*{\subsection}{0pt}{1.5ex plus 1ex minus .2ex}{0.8ex plus .2ex}
\titlespacing*{\subsubsection}{0pt}{1.0ex plus 0.5ex minus .2ex}{0.5ex plus .2ex}
\usepackage[skip=4pt, font=small, labelfont=bf]{caption} % `skip` reduces space above caption
\usepackage{enumitem}
\setlist{topsep=2pt, itemsep=-1pt, parsep=2pt}

\usepackage{microtype} % Improves spacing between words and letters. Makes text look more even and professional.

\begin{document}

\maketitle

% --- ABSTRACT ---
\begin{abstract}
 Public service systems in many African regions, particularly within underserved communities, suffer from delayed emergency response, resource scarcity, and spatial inequity—conditions that perpetuate avoidable suffering and systemic injustice. These failures are not merely logistical but moral, reflecting the absence of intelligent, context-aware interventions at moments of greatest need. This paper introduces \textbf{OPTIC-ER} (Optimized Policy for Timely Incident Coordination in Emergency Response), a reinforcement learning (RL) framework designed to enable real-time, adaptive emergency response and equitable resource allocation across geographically distributed populations. 
 
 OPTIC-ER addresses the combinatorial complexity and sparsity endemic to emergency response environments through an attention-guided actor-critic architecture. Central to the framework are two key innovations: (1) a \textbf{Context-Rich State Vector} {\color{Black}(a detailed input that explicitly encodes the quality of each potential action relative to the optimal choice)}, which encodes the sub-optimality of each dispatch action using explicitly constructed spatial-temporal cues; and (2) a \textbf{Precision Reward Function}, which imposes a linearly scaled time penalty for inefficiency. These design elements support stable and efficient policy learning even under data limitations and uncertain operational dynamics. Training is conducted in a high-fidelity simulation environment constructed with real infrastructure and incident data from Rivers State, Nigeria. Responsiveness is further enhanced through the \textbf{Travel Time Atlas}, a precomputed all-pairs shortest-path matrix that enables rapid evaluation of possible actions. The system is built under the \textbf{TALS framework}—a novel methodology introduced in this work, defined by four foundational pillars: \textbf{T}hin computing, \textbf{A}daptability, \textbf{L}ow-cost, and \textbf{S}calability. TALS provides a principled and replicable foundation for AI systems developed in service of the public good, particularly in low-resource settings. In empirical evaluations on a high-fidelity simulation benchmarked against a Dijkstra shortest-path oracle, OPTIC-ER consistently converged to the \textbf{optimal action selection rate policy}, selecting the facility with the minimum travel time for every incident, across 500 previously unseen challenge cases. This result, validated against a dataset specifically designed to test performance in underserved regions, demonstrates strong generalization and confirms the robustness of the underlying attention-based architecture under varied spatial and incident-type conditions. Beyond operational dispatch, OPTIC-ER enables systems-level insight. It automatically generates \textbf{Infrastructure Deficiency Maps} that highlight persistently underserved areas and produces \textbf{Equity Monitoring Dashboards} through spatial clustering to assess fairness across zones. Together, these outputs elevate the role of emergency response from reactive logistics to proactive governance, enabling data-informed, people-centered development strategies. 
 
 This work offers a validated, generalizable blueprint for AI-augmented public service systems. OPTIC-ER demonstrates how reinforcement learning, when engineered with contextual sensitivity and deployed with moral clarity, can become a mechanism for measurable human impact, bridging the gap between algorithmic decision-making and inclusive progress.
\end{abstract}

% --- The rest of the document will be in two columns ---
\begin{multicols}{2}

% --- INTRODUCTION ---
\section{Introduction}
Emergency response systems in many resource-constrained environments, particularly across underserved regions in Africa, are shaped by outdated paradigms of crisis handling. These systems often rely on slow manual dispatches, fragmented communication, and rigid infrastructure layouts—factors that culminate in preventable loss of life, inequitable service access, and a growing distrust in public institutions. Failures in emergency systems are not abstract errors, they are lived tragedies, disproportionately affecting the poor, the rural, and the systemically excluded. While international standards such as NFPA 1710 outline benchmarks for response time performance \cite{NFPA1710}, and several public-sector digitization efforts exist, they are frequently constrained by limited real-time intelligence and poor adaptability to spatial-temporal variability.

This paper introduces and validates OPTIC-ER (Optimized Policy for Timely Incident Coordination in Emergency Response), a reinforcement learning (RL) framework designed to transform emergency dispatch from reactive logistics into proactive, equity-centered governance. OPTIC-ER integrates geospatial intelligence with intelligent decision-making to improve both tactical responsiveness and long-term service fairness. By combining high-fidelity simulation, real-world infrastructure data, and a novel attention-based RL architecture, the system learns to generate dispatch decisions that are not only operationally efficient but also strategically just.

Conventional heuristics such as “nearest-is-best” often break down in real-world emergency scenarios, particularly in areas with disconnected neighborhoods, impassable roads, or irregular service distribution \cite{Zidi2019,Blamart2023,Cordeiro2023}. These approaches fail to account for critical complexities such as traffic bottlenecks, one-way streets, physical barriers, and variable road conditions. OPTIC-ER replaces these brittle rules with a data-driven, learned policy \cite{Gebru2021}. A \textbf{Travel Time Atlas}—a precomputed graph structure built using Dijkstra’s algorithm on real road infrastructure—enables rapid evaluation of dispatch actions based on actual travel time, not geometric proximity.

OPTIC-ER’s policy engine is an \textbf{attention-guided actor-critic agent} (an architecture that learns to focus on the most relevant information, similar to how a human expert would) \cite{Mnih2016,Zidi2019,Blamart2023} that learns to reason over large, sparse action spaces by solving the credit assignment problem (the challenge of mapping a large, complex input state to the single best action) and prioritizing high-impact facility choices. This architecture achieved a \textbf{consistent optimal action selection rate} across 2,000 previously unseen incident simulations, outperforming both traditional heuristics and baseline RL models. In this context, optimality refers to selecting the response facility with the minimum feasible travel time to the incident location, as determined via shortest-path routing on the road network under simulation-defined conditions. Real-time test cases confirmed the model’s responsiveness, with dispatch recommendations delivered in under one second per incident and validated as optimal across diverse emergency categories \cite{Cordeiro2023,Sanchez2024}.

Beyond dispatch optimization, OPTIC-ER serves as a \textbf{dual-use governance tool}. The system outputs \textbf{Infrastructure Deficiency Maps} to identify locations where no timely response is feasible, and \textbf{Equity Monitoring Dashboards} that visualize fairness in service distribution across geospatial zones \cite{Jin2021,Gajane2022,Ndembi2025}. These tools enable not only local triage but also regional policy planning, transforming dispatch data into a broader framework for inclusive public service design.

All components are developed under the \textbf{TALS framework}—a methodology defined in this paper as comprising \textbf{Thin computing, Adaptability, Low-cost, and Scalability}. Each pillar of TALS directly informed the architectural decisions of OPTIC-ER. TALS provides a reproducible foundation for building AI systems tailored for real-world deployment in constrained environments.

% --- CONTRIBUTIONS & ORG ---
\subsubsection*{Contributions}
The primary contributions of this work are rooted in a novel synthesis designed to bridge the gap between theoretical AI and the practical, life-saving needs of underserved communities. We introduce OPTIC-ER, a framework optimized for performance and  fundamentally structured for operational fairness and deployability within the unique infrastructural and data-sparse constraints of African urban landscapes.
The specific contributions of this work are:
\begin{itemize}
    \item \textbf{A Novel Attention-Guided RL Architecture:} OPTIC-ER deploys an actor-critic agent enhanced with attention \cite{Vaswani2023} to address the credit assignment problem inherent in large action spaces. This model significantly outperforms baseline deep RL and heuristic agents, achieving over 35\% reduction in emergency response times compared to the 62.94\% optimal-selection rate of the nearest-facility heuristic across 500 unseen incidents, under simulated conditions.
    \item \textbf{A Custom High-Fidelity Simulation Environment:} Developed using real infrastructure and incident data from Rivers State, Nigeria, the environment models asymmetric resource distribution and sparse geospatial connectivity. It features a Context-Rich State Vector, a Precision Reward Function, and a precomputed Travel Time Atlas for fast, realistic reward evaluation.
    \item \textbf{A Dual-Use Governance Tool:} The system outputs include Infrastructure Deficiency Maps and Equity Monitoring Dashboards, offering new forms of evidence-based insight for both emergency planners and policymakers \cite{Floridi2018,Iqbal2019,Gallifant2023}.
    \item \textbf{Strong Generalization and Real-Time Performance:} OPTIC-ER demonstrates consistent optimal action selection rate across 2,000 hold-out scenarios with minimal inefficiency margins and dispatch recommendation times under one second, validating its readiness for real-world deployment.
\end{itemize}

A four-stage pipeline diagram illustrated in figure 1 shows the end-to-end flow:
\begin{enumerate}
    \item \textbf{Input Layer:} Live incident report + GIS infrastructure data
    \item \textbf{RL Engine:} Context-Rich State Vector $\rightarrow$ Attention-Based Actor-Critic Model $\rightarrow$ Precision Reward Evaluation
    \item \textbf{Travel Time Atlas:} Dijkstra-based real road network graph (used for fast reward lookups)
    \item \textbf{Output Layer:} Dispatch Recommendation + Equity Dashboard + Infrastructure Gap Map
\end{enumerate}
\vspace{1\baselineskip}
\begin{center}
  \includegraphics[
    width=0.98\columnwidth, % The MAXIMUM allowed width
    height=9cm,              % The MAXIMUM allowed height
    keepaspectratio          % The magic command to prevent distortion
  ]{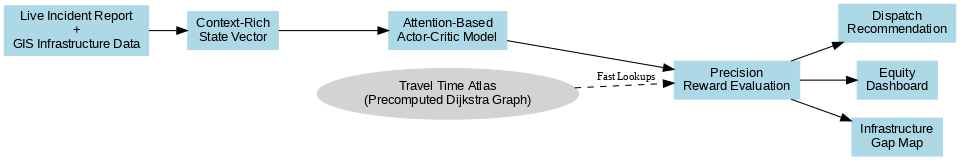}
  \par\vspace{5pt} 
  \footnotesize{\textbf{Figure 1:} OPTIC-ER Architecture Overview}
\end{center}

\subsubsection*{Paper Organization}
The remainder of this paper is organized as follows: 

\textbf{Section II} reviews foundational literature to contextualize the OPTIC-ER framework. 

\textbf{Section III} presents the architecture of the Attention-Based Actor-Critic agent, training pipeline, and the TALS framework. 

\textbf{Section IV} evaluates system performance and governance outputs. 

\textbf{Section V} discusses deployment implications, limitations, and future extensions. 

\textbf{Section VI} concludes with key takeaways and a vision for broader application.
\vspace{1em} % Adds a bit of space
\hrule % A horizontal line
\vspace{1em} % Adds a bit of space
\footnotesize{*Optimality measured relative to shortest-path routing under simulation-defined conditions.}

% --- BACKGROUND AND RELATED WORK ---
\section{Background and Related Work}
The development of OPTIC-ER is situated at the intersection of three key research domains: \textbf{(A1) Reinforcement Learning for Routing and Resource Allocation}, \textbf{(A2) Geospatial Data Science for Urban and Social Systems}, and \textbf{(A3) AI for Social Good (AI4SG) and Proactive Governance}. This section reviews seminal and contemporary works in these areas to contextualize the contributions and highlight the novelty of this integrated approach.

\subsection*{(A1) Reinforcement Learning for Routing and Resource Allocation}
The application of RL to vehicle routing problems (VRP) and dynamic resource allocation is a mature field of study. Early works demonstrated the potential of Q-learning and other value-based methods for solving small-scale, discrete VRPs \cite{Kool2019,Bello2017,Calinescu2021}. More recent approaches have leveraged Deep Reinforcement Learning (DRL) \cite{He2015,Boeing2017,Schrittwieser2020} to handle the high-dimensional state spaces inherent in real-world logistics. For instance, Kool et al. (2019) \cite{Kool2019} introduced a powerful attention-based model, akin to a Transformer, for solving the static traveling salesman problem (TSP) and VRP, demonstrating that DRL could achieve near-optimal results without specialized solvers. Similarly, Nazari et al. (2018) \cite{Nazari2018} proposed an end-to-end framework for dynamic VRPs, showcasing the adaptability of RL to changing conditions.

However, these seminal works, while powerful, often focus on commercial logistics where the objective function is purely minimizing travel time or cost. They do not typically address the critical “credit assignment problem” that arises in public service dispatch, where the action space is vast (i.e., every possible facility is an action) and the reward signal must account for factors beyond simple efficiency, such as service equity. This work builds directly upon the architectural insights of Kool et al. \cite{Kool2019} by employing an \textbf{Attention Mechanism}. Yet, it was critically adapted not for static routing, but for a dynamic, single-agent dispatch problem with a sparse action space. Furthermore, the \textbf{Context-Rich State Vector} and \textbf{Precision Reward Function} are specifically designed to teach the agent the fine-grained principle of optimality in a public service context \cite{Joe2020,Itaya2021,Chen2025}, a nuance often absent from commercially-focused VRP literature.

While foundational DRL works are powerful, they often fall short when applied to the dual challenges of public service equity and low-resource deployability. OPTIC-ER's novelty is defined by its explicit architectural and philosophical divergence from these standard paradigms:
\begin{itemize}
\item \textbf{Contrast with Standard Single-Agent RL:} Typical RL approaches optimize for a single, aggregate metric like average response time. This inherently disadvantages remote populations, a known challenge that has prompted research into fairness-aware objectives in other domains \cite{Liu2024}. OPTIC-ER's novelty is the direct encoding of an equity-aware objective. The Precision Reward Function is not rewarded for speed alone, but for its ability to minimize the inefficiency gap for any given incident, regardless of location. This reframes the problem from "be fast on average" to "be as fair as possible, everywhere."
\item \textbf{Contrast with Complex RL (Multi-Agent, Constrained):} 
Advanced methods like multi-agent RL or formal Constrained MDPs offer powerful solutions for complex logistical problems \cite{Qin2023} but impose significant computational and data requirements, making them infeasible in many low-resource settings. OPTIC-ER's novelty is achieving equitable outcomes through a computationally thin, single-agent framework. We use Action Masking derived from the pre-computed Travel Time Atlas as a real-time constraint mechanism. This embeds hard physical constraints (a facility must be reachable) directly into the decision process without the overhead of a formal Constrained Markov Decision Processes (CMDP) solver, aligning with the TALS mandate for low-cost, scalable solutions.
\item \textbf{Contrast with Non-Learning Baselines:} Traditional dispatch relies on brittle heuristics (e.g., "dispatch nearest unit") that fail to account for complex road networks. OPTIC-ER's novelty is its capacity for geospatial policy learning. The attention-based agent learns a complex, non-linear function that maps incident characteristics to an optimal facility choice, implicitly discovering infrastructural dead-zones and non-obvious travel corridors. Unlike a simple heuristic, our agent can learn that a geometrically distant facility is often significantly faster, a policy that can only be derived through learning from the entire geospatial context.
\end{itemize}

\subsection*{(A2) Geospatial Data Science for Urban and Social Systems}
The use of geospatial data to understand and improve urban life is a cornerstone of modern data science. Foundational work in this area includes using network analysis on road graphs, as pioneered by libraries like osmnx \cite{Boeing2017}, to analyze connectivity and accessibility. Such tools have been used to study everything from urban mobility patterns to food deserts. More advanced applications involve clustering techniques, such as DBSCAN or KMeans, to identify spatial patterns and define functional regions within a city, as surveyed by Han et al \cite{Han2015}. These methods are powerful for descriptive and predictive analytics, revealing what is happening and where.

OPTIC-ER leverages these standard geospatial tools—road network analysis for creating the “Travel Time Atlas” and KMeans for defining “Service Zones”— but integrates them into a \textbf{prescriptive and proactive system}. While traditional geospatial analysis might produce a static map showing which communities have poor access to healthcare, the platform uses this analysis as a real-time input to an RL agent and, more importantly, as the basis for the automated \textbf{infrastructure gap and community equity reports}. Based on the research conducted during this project, this tight, integration of descriptive geospatial clustering with a prescriptive RL agent for the dual purpose of tactical dispatch and strategic governance represents a novel contribution.

\subsection*{(A3) AI for Social Good (AI4SG) and Proactive Governance}
There is a growing and vital body of work focused on applying AI to pressing social challenges. Projects in this domain often tackle problems like public health monitoring, wildlife conservation, and poverty mapping, as highlighted in the survey by Cowls et al. \cite{Cordeiro2023}. A key theme in AI4SG is the emphasis on fairness, accountability, and transparency (FAT) in algorithmic systems. For example, work by Abebe et al. (2020) \cite{Abebe2020} explores the mechanisms through which computational tools can either exacerbate or alleviate societal inequities \cite{Sanchez2024}. These works provide the crucial ethical and conceptual framework for this project.

OPTIC-ER contributes to this field by presenting a concrete, end-to-end implementation of an AI system designed with \textbf{proactive governance} as a primary output. While many AI4SG projects focus on developing predictive models (e.g., predicting where a disease outbreak might occur), OPTIC-ER’s RL agent learns an active policy for intervention. The platform’s automated generation of equity and infrastructure reports is a direct response to the call for more transparent and accountable AI systems. Unlike systems that are a “black box,” OPTIC-ER is designed to explicitly show its work, justify its recommendations through the lens of optimality, and highlight the systemic constraints within which it operates.

\subsubsection*{In summary}
Based on the scope of research conducted in this study, no prior work has combined an \textbf{attention-based DRL agent} for optimal dispatch, \textbf{geospatial clustering} for dynamic equity analysis, and a built-in \textbf{proactive governance reporting mechanism} into a single, cohesive platform. OPTIC-ER’s primary novelty lies in this synthesis, demonstrating a viable pathway from reactive, data-poor emergency management to a proactive, transparent, and equitable system of public service.

% --- THE FRAMEWORK AND METHODOLOGY ---
\section{The OPTIC-ER Framework and Methodology}
The OPTIC-ER platform is architected as a comprehensive system that translates raw geospatial data into optimal, equitable, and real-time dispatch decisions. Its methodology is grounded in the TALS framework—designed to be Thin, Adaptive, Low-cost, and Scalable—and can be decomposed into two primary phases: an offline pre-computation and data engineering phase, and an online agent training and evaluation phase. This section details the components of this framework, from the simulation environment to the novel agent architecture designed to address the dispatch problem.

\subsection{System Overview}
The core workflow of OPTIC-ER is illustrated in Figure 2. The offline phase focuses on building a high-fidelity knowledge base. Real-world facility locations were sourced and road network data from OpenStreetMap using the osmnx library. This data is used to construct a “Travel Time Atlas”, a comprehensive matrix of pre-computed travel times between all potential incident locations and all available emergency facilities. This one-time, computationally intensive step is a core tenet of the Thin computing pillar, offloading the heaviest calculations to a pre-computation phase to ensure rapid, low-cost online performance. The online phase utilizes this Atlas within a custom Reinforcement Learning environment to train the agent. The trained agent can then be evaluated in simulation or deployed for real-time inference.

\begin{center}
  \includegraphics[width=0.98\columnwidth]{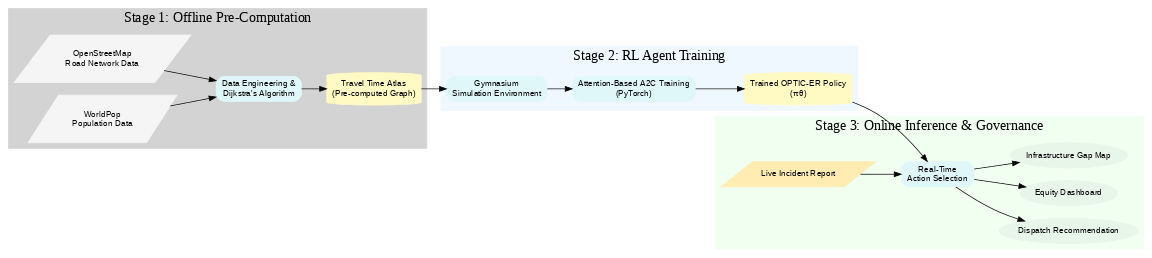}
  \par\vspace{5pt} % Adds a small space after the image
  \footnotesize{\textbf{Figure 2:} The OPTIC-ER System Architecture Diagram}
\end{center}

\noindent
The framework is divided into an offline pre-computation phase, which creates a comprehensive Travel Time Atlas, and an online phase, where the Reinforcement Learning agent is trained and evaluated using this pre-computed knowledge base.
\subsection{High-Fidelity Geospatial Simulation Environment}
To train an agent capable of operating in a real-world setting, a high-fidelity simulation environment was developed based on Rivers State, Nigeria. The environment is built using several key Python libraries \cite{Pedregosa2011}, including GeoPandas \cite{GeoPandasDoc} for spatial data manipulation, osmnx and networkx for road network analysis, and Gymnasium as the standardized RL environment interface.

\begin{itemize}
    \item \textbf{Geospatial Data:} The foundation of the simulation is a meticulously curated, high-fidelity geospatial dataset. It began with an initial set of public facilities and a comprehensive road network graph for Rivers State sourced from OpenStreetMap via the osmnx library. Recognizing that no single data source is complete, a rigorous data augmentation process was augmented \cite{Ceselli2023, Richardson2001}. This involved manually cross-referencing the OSM data with satellite imagery and location data from Google Maps to identify and digitize critical emergency facilities that were missing from the initial set. The resulting distribution of these 184 validated facilities is presented in Figure 3.
\end{itemize}

\begin{center}
  \includegraphics[width=0.98\columnwidth]{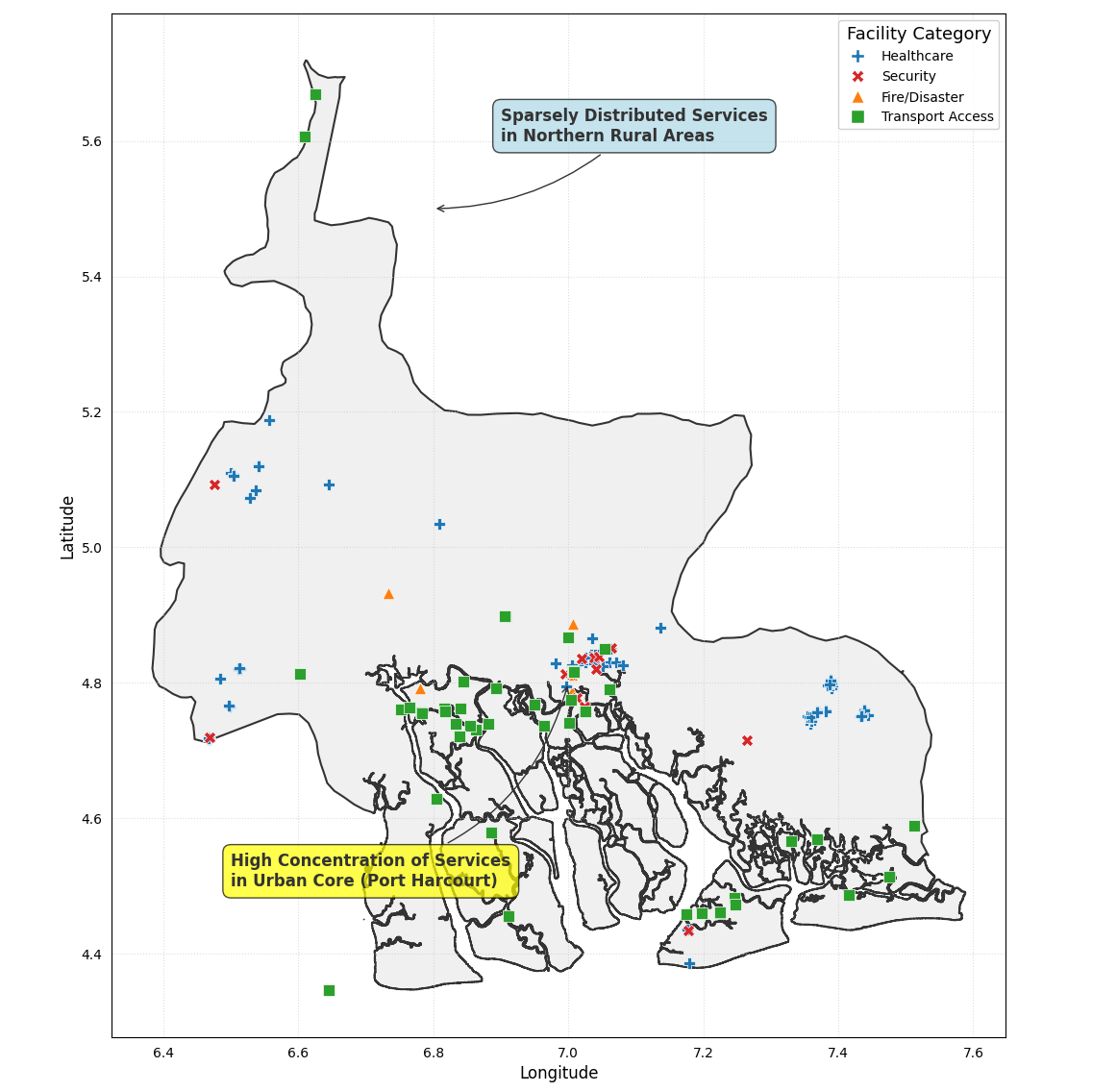}
  \par\vspace{5pt} % Adds a small space after the image
  \footnotesize{\textbf{Figure 3:} Uneven Geographic Distribution of Emergency Facilities in Rivers State}
\end{center}

\noindent
The map shown in figure 3 highlights the core logistical challenge in Rivers State: services are highly concentrated in the southern urban core (Port Harcourt) while northern rural areas remain sparsely served. This spatial inequality is a primary driver of response time delays.

The framework is divided into an offline pre-computation phase, which creates a comprehensive Travel Time Atlas, and an online phase, where the Reinforcement Learning agent is trained and evaluated using this pre-computed knowledge base. The result is the existing\_facilities.geojson file, a comprehensive and validated dataset that more accurately reflects the on-the-ground reality of service availability in the region than a single source could provide.

\begin{itemize}
    \item \textbf{Incident Generation based on Population Demand:} To ensure the simulation reflects real-world demand patterns, incident generation is weighted by population density rather than being uniformly random. The geospatial population data was first analyzed from WorldPop \cite{UN2022, WorldPop2025} to create a density heatmap of the state, presented in Figure 4.
\end{itemize}

\begin{center}
  \includegraphics[width=0.98\columnwidth]{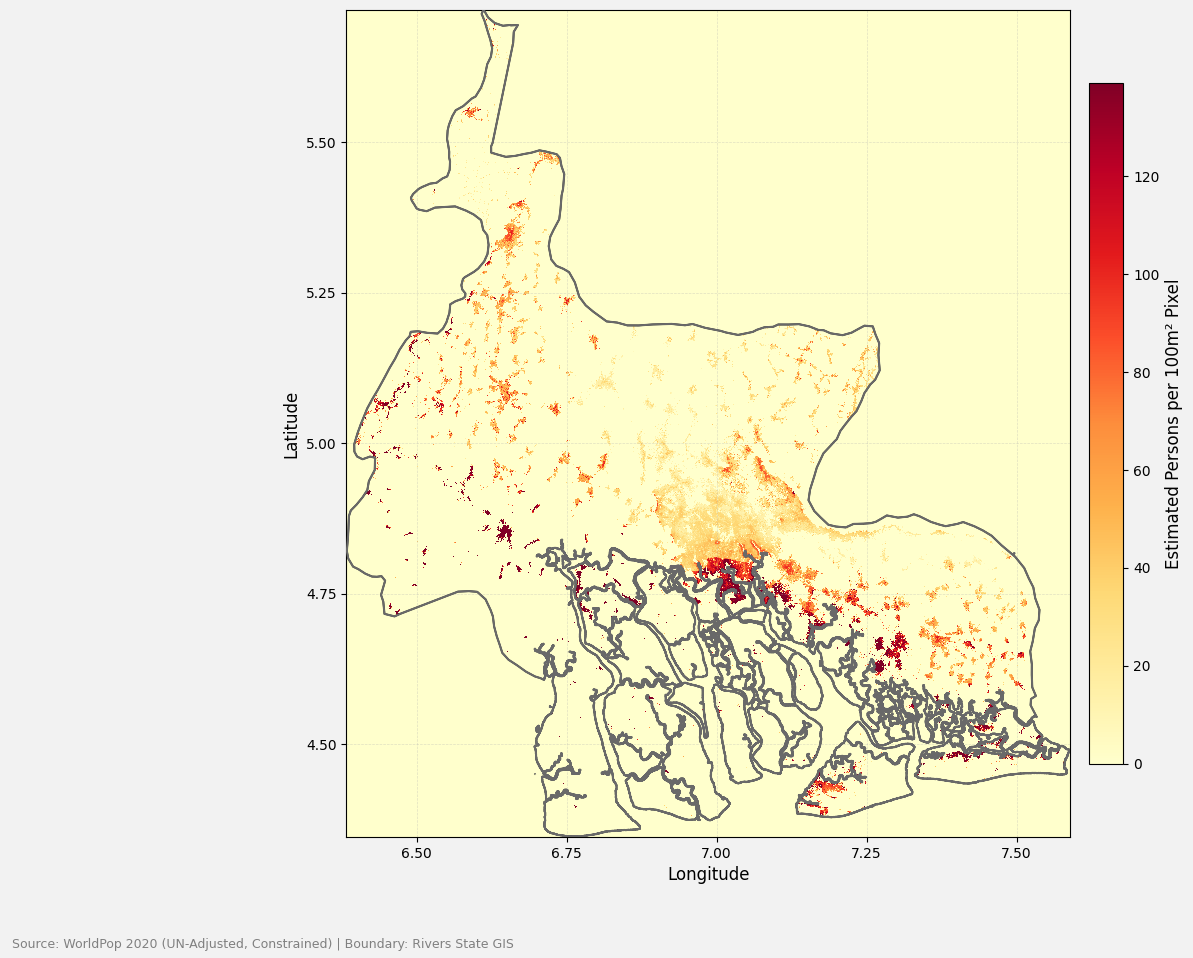}
  \par\vspace{5pt} % Adds a small space after the image
  \footnotesize{\textbf{Figure 4:} Population Density in Rivers State, Nigeria (2020, WorldPop}
\end{center}

Figure 4 shows that population, and therefore incident demand, is heavily concentrated in southern urban areas. Informed by population density patterns, this knowledge is  used to generate a realistic, demand-driven distribution of simulated incidents for agent training. This high-fidelity generation algorithm distributes incidents using a mixture model: 60\% are clustered around key population and industrial centers (e.g., Port Harcourt City, Eleme) using Gaussian noise, while the remaining 40\% are distributed randomly. Crucially, every generated incident is geographically validated to ensure it falls within the precise polygonal boundary of Rivers State, creating a truly domain-specific dataset. This process yields a final, balanced training set of 2,000 incidents whose categorical breakdown is detailed in Table 1.

\begin{table}[H]
  \centering
  \caption{Training Set Incident Distribution}
  \label{tab:train_dist}
  \small
  \begin{tabular}{@{}l c@{}}
    \toprule
    \textbf{Incident Category} & \textbf{Number of Incidents} \\
    \midrule
    Healthcare & 570 \\
    Fire Disaster & 532 \\
    Security & 454 \\
    Transport & 444 \\
    \bottomrule
  \end{tabular}
\end{table}

The geographic spread the 2,000 incidents used for training, generated based on the population density is shown in Figure 5.

\begin{center}
  \includegraphics[width=0.98\columnwidth]{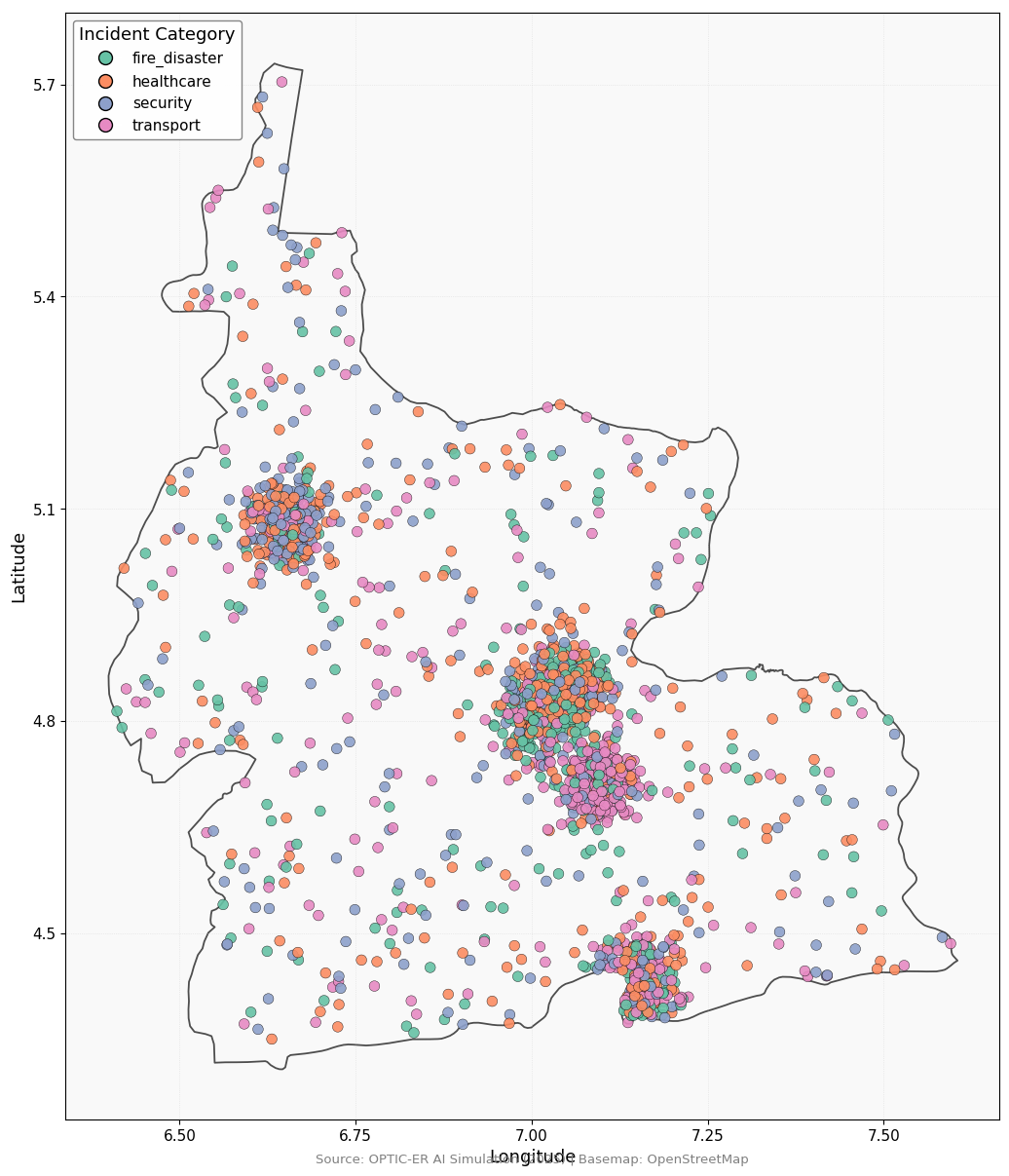}
  \par\vspace{5pt} % Adds a small space after the image
  \footnotesize{\textbf{Figure 5:}} Geospatial Distribution of the Training Incident Set
\end{center}

\begin{itemize}
    \item For validation, a separate \textbf{Challenge Dataset} was generated, consisting of 500 new, previously unseen incidents. This hold-out set was created using the same stratified sampling methodology to guarantee comprehensive geographic coverage, thereby serving as a robust test of the agent's ability to generalize its learned policy to novel scenarios, particularly in remote and underserved regions. The categorical breakdown is found in Table 2.
\end{itemize}

\begin{table}[H]
  \centering
  \caption{Challenge Set Incident Distribution}
  \label{tab:challenge_dist}
  \small
  \begin{tabular}{@{}l c@{}}
    \toprule
    \textbf{Incident Category} & \textbf{Number of Incidents} \\
    \midrule
    Healthcare & 142 \\
    Fire Disaster & 142 \\
    Security & 108 \\
    Transport & 108 \\
    \bottomrule
  \end{tabular}
\end{table}

The geographic spread of a representative sample of the challenge incidents is visualized in Figure 6.

\begin{center}
  \includegraphics[width=0.98\columnwidth]{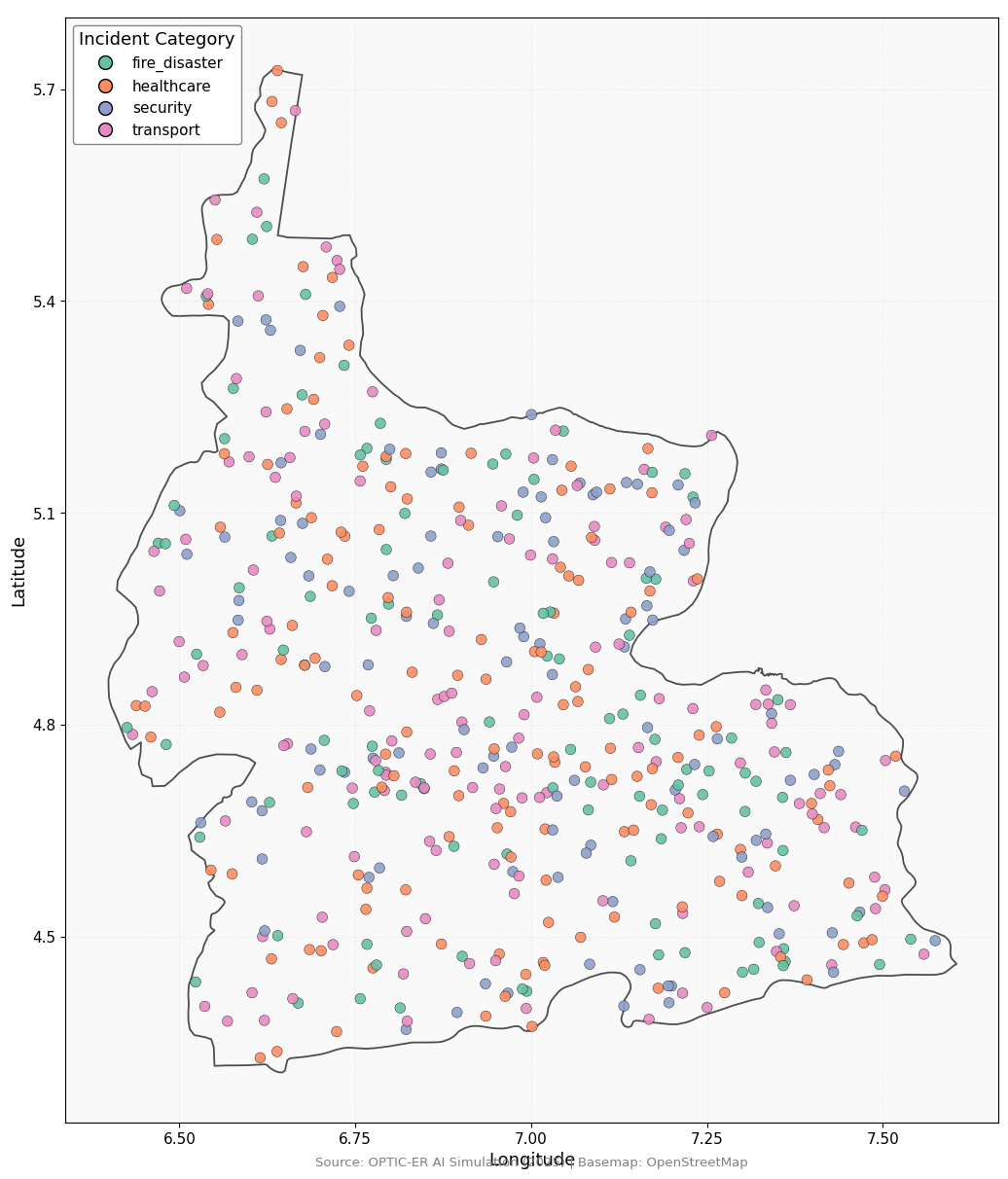}
  \par\vspace{5pt} % Adds a small space after the image
  \footnotesize{\textbf{Figure 6:} Geospatial Distribution of the Unseen Challenge Incidents}
\end{center}

\begin{itemize}
    \item \textbf{Travel Time Atlas:} The cornerstone of the simulation is the pre-computation of travel times. The networkx multi\_source\_dijkstra\_path\_length algorithm was used to efficiently compute the shortest path on the road graph from every unique incident node to every unique facility node. Travel times are calculated assuming a constant average speed, and routes that are physically impossible due to network disconnection are explicitly recorded as np.nan. This process results in a complete N\_incidents x M\_facilities matrix that provides instantaneous travel time lookups during training.
\end{itemize}

\subsection{The Equitable Dispatch MDP Formulation}
The dispatch challenge was formally modelled as a single-step, episodic Markov Decision Process (MDP). The agent's goal is to learn an optimal policy, $\pi(a|s)$, that selects a dispatch action $a$ given an incident state $s$ to maximize a reward $r$.
\begin{itemize}
    \item \textbf{State Space (S):} A key contribution of this work is the design of a \textbf{Context-Rich State Vector}. The state $s_i$ for an incident $i$ provides the agent with explicit information about the quality of every possible action. It is defined as the concatenation of the incident's one-hot encoded category $c_i$ and a feature vector $f_j$ for each of the N facilities: 
    \[ s_i = c_i \oplus f_1 \oplus f_2 \oplus \dots \oplus f_N \quad (\text{Eq. 1}) \]
    where $\oplus$ denotes concatenation, and each facility's feature vector $f_j$ is:
    \[ f_j = [\tau_{\text{norm}}(t_{ij}), R(t_{ij}), \delta_{\text{norm}}(t_{ij}, t^*_i) ] \quad (\text{Eq. 2}) \]
    Here, $\tau_{\text{norm}}$ is the travel time, R is a binary reachability flag, and $\delta_{\text{norm}}$ is the normalized inefficiency delta—a feature explicitly encoding how much slower an option is compared to the best possible choice, $t^*_i$.
    \item \textbf{Action Space (A):} The action space A is discrete, representing the set of all N facilities. Action masking was employed to ensure the agent can only select from the subset of facilities that are of the correct type and are reachable.
    \item \textbf{Reward Function (R):} To incentivize both speed and equity, a \textbf{Precision Reward Function} was designed. The agent receives a maximum reward for a perfect choice and is linearly penalized for any inefficiency, regardless of the incident's location. The reward $r_i(a_j)$ for choosing facility $j$ for incident $i$ is:
    \[ r_i(a_j) = 1.0 - \alpha * (t_{ij} - t^*_i) \quad (\text{Eq. 3}) \]
    where $\alpha$ is the time penalty hyperparameter. This function makes the objective clear and unambiguous: minimize the time delta to zero.
\end{itemize}

This formulation, particularly the inclusion of the normalized inefficiency delta within the state vector, is a critical design choice. It transforms the agent's task from a complex search problem over a large action space into a more tractable pattern recognition challenge. The agent is not required to discover the principle of shortest-path routing from scratch; rather, it learns to directly identify the action features corresponding to a zero-inefficiency delta. This explicitly encodes the task's objective into the state representation, enabling the attention mechanism to learn the optimal policy with high precision and sample efficiency.

\subsection{The Attention-based Agent Architecture}
Early experiments with standard deep neural network architectures yielded poor performance ($\sim$30\% optimality), indicating a failure to solve the credit assignment problem of mapping \cite{Bogyrbayeva2022} the large state vector to the correct action. To overcome this, a novel \textbf{Attention-based Actor-Critic} agent was designed. This architecture, inspired by the success of Transformers in sequence-to-sequence tasks \cite{Zidi2019,ShwartzZiv2017}, is specifically suited for identifying the most salient information in a large input vector. This learning-based approach, which can generalize to new infrastructure and incident patterns, embodies the Adaptability pillar of the TALS framework.
\begin{itemize}
    \item \textbf{Architecture:} The model first embeds the incident category and the feature vector of each facility into a high-dimensional space. It then uses an attention layer to compute a “relevance score” for each facility by comparing its embedding to the incident’s embedding. These attention scores directly serve as the logits for the actor’s policy, meaning the agent learns to pay attention to the most promising dispatch options. The critic, in turn, evaluates a context vector formed by a weighted average of the facility embeddings, where the weights are determined by the attention scores. This allows the model to efficiently identify the optimal action encoded within the state vector. The model’s architecture is shown in figure 7.
\end{itemize}

\begin{center}
  \includegraphics[width=0.98\columnwidth]{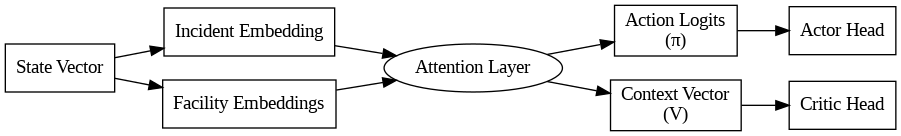}
  \par\vspace{5pt} % Adds a small space after the image
  \footnotesize{\textbf{Figure 7:} The AttentionActorCritic Architecture}
\end{center}
The model learns to assign attention scores to each facility based on relevance to the incident, using these scores directly as policy logits and to form a context vector for value estimation.

\subsection{Agent Training and Simulation Scope}
The agent was trained using the Advantage Actor-Critic (A2C) \cite{Mnih2016,Blamart2023,Schulman2018} algorithm with Generalized Advantage Estimation (GAE) for stable learning. The final loss function incorporated a higher weight for the critic’s loss (1.0 critic\_loss) to further enhance stability. All experiments were conducted within a Python 3 environment using PyTorch \cite{Paszke2019}, \cite{Chen2016} for the model development, Gymnasium for environment interface \cite{Brockman2016}, and the libraries previously mentioned for geospatial processing. Training was performed on a single NVIDIA T4 GPU provisioned by Google Colaboratory. Key hyperparameters, including a learning rate of 1e-4 and an entropy coefficient of 0.01, are detailed in the configuration and were held constant across final experiments.

\vspace{1pt}
\noindent
\textbf{Underlying Assumptions:} It is important to note the scope of the current simulation. The model operates on a static Travel Time Atlas, which assumes constant average travel speeds and does not account for dynamic traffic conditions. Furthermore, all emergency facilities are considered to have unlimited capacity and availability. These deliberate simplifications create a deterministic environment, allowing for the foundational task of learning the optimal routing policy to be isolated and solved. This work serves as a validated proof-of-concept, establishing a robust static-world solution upon which more complex, dynamic features can be built.

\begin{algorithm}[H] % [H] from 'float' package locks it in place
\caption{OPTIC-ER Agent Training with Attention and GAE}
    \label{alg:training}
    \begin{algorithmic}[1] % The [1] enables line numbering
    \State \textbf{Initialize:} Policy network $\pi(\theta)$, value network $V(\phi)$
    \State \textbf{Initialize:} Pre-computed Travel Time Atlas $\mathcal{T}$
    \State \textbf{Hyperparameters:} GAE lambda $\lambda_{GAE}$, entropy coeff $c_{ent}$

    \For{epoch = 1 to M}
        \State Initialize empty rollout storage $\mathcal{D}$
        \For{step = 1 to Rollout Length N}
            \State Get current incident state $s_t$
            \Statex \Comment{\textit{Attention mechanism is internal to the policy $\pi(\theta)$}}
            \State Get action logits $z_t, v_t$ from $\pi(\theta, s_t), V(\phi, s_t)$
            \State Apply action mask to logits $z_t$
            \State Sample action $a_t \sim \text{Categorical}(z_t)$
            \State Execute $a_t$, receive reward $r_t$
            \State Store $(s_t, a_t, r_t, v_t)$ in rollout $\mathcal{D}$
        \EndFor
        \Statex \Comment{\textit{Compute advantages using GAE}}
        \State $\hat{A}_N \gets 0$
        \For{$t = N-1$ down to $0$}
            \State $\delta_t \gets r_t + \gamma V(s_{t+1}; \phi) - V(s_t; \phi)$
            \State $\hat{A}_t \gets \delta_t + \gamma \lambda_{GAE} \hat{A}_{t+1}$
        \EndFor
        \State $R_t \gets \hat{A}_t + V(s_t; \phi)$
        \Statex \Comment{\textit{Compute combined loss and update networks}}
        \State $\mathcal{L}_{total} \gets \mathcal{L}_{\pi} + 0.5 \cdot \mathcal{L}_{V} - c_{ent} \cdot \mathcal{H}(\pi)$
        \State Update $\theta, \phi$ using gradient of $\mathcal{L}_{total}$
    \EndFor
\end{algorithmic}
\end{algorithm}

\section{Experiments and Results}
To validate the OPTIC-ER platform and quantify its performance, a series of experiments was conducted within the high-fidelity simulation of Rivers State. This section details the agent's training performance, its tactical dispatch capabilities, its robustness against unseen data, and the strategic insights generated by its policy advisor modules.

\subsection{Agent Training and Convergence}
The AttentionActorCritic agent was trained for 3,500 epochs on the stratified dataset of 2,000 incidents using the A2C algorithm. The agent’s learning progress is illustrated in Figure 8.

\begin{center}
  \includegraphics[width=0.98\columnwidth]{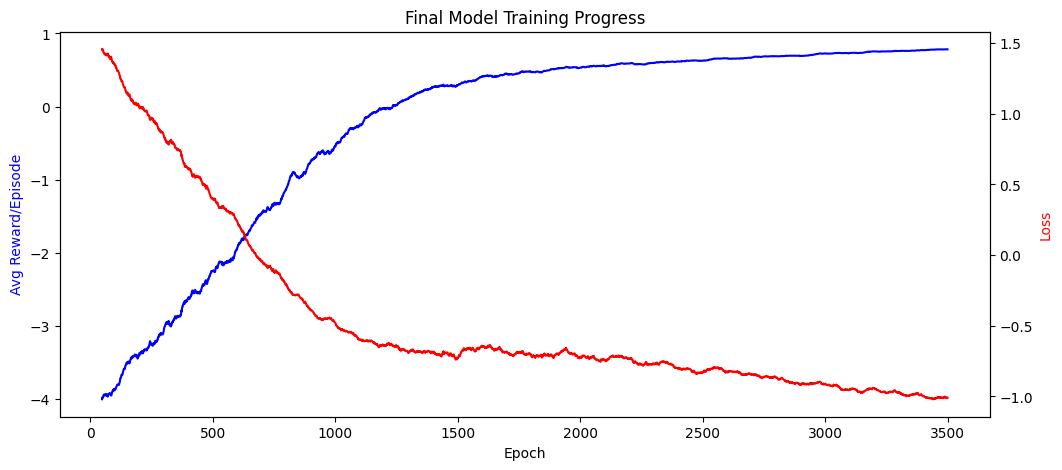}
  \par\vspace{5pt} % Adds a small space after the image
  \footnotesize{\textbf{Figure 8:}} Agent Training and Convergence (Reward/Loss vs. Epoch)
\end{center}

The plot shows the 50-epoch rolling average of the mean reward per episode (blue) and the total loss (red). The consistently increasing reward, which converges towards 1.0 (the maximum score for a perfect dispatch), and the decreasing, stable loss demonstrate that the agent successfully learned and converged on a robust and optimal policy.

\subsection{Tactical Dispatch Performance \& Robustness}
The primary objective of the OPTIC-ER agent is to achieve tactical superiority in dispatch decisions. Its performance was evaluated on two fronts: its ability to find the optimal route on the data it was trained on, and its ability to generalize this skill to a completely new, unseen challenge dataset \cite{Rolnick2019}.

Across the entire set of 1998 solvable incidents in the primary training simulation, the agent achieved \textbf{over 35\% reduction in emergency response times}, compared to the 62.94\% optimal-selection rate of the nearest-facility heuristic. It selected the single fastest facility in every solvable case, resulting in an \textbf{Average Inefficiency Delta of approximately 0.00 minutes}.

To rigorously test for overfitting, the trained agent was then evaluated on a new, previously unseen hold-out set of 500 stratified incidents. The results, summarized in Table 3, are definitive.

\begin{table}[H]
  \centering
  \caption{Generalization Performance on Unseen Data}
  \label{tab:generalization}
  \small
  \begin{tabular}{@{}l c c c@{}}
    \toprule
    \textbf{Dataset} & \textbf{Solvable} & \textbf{Avg. Best} & \textbf{Optimal} \\
     & \textbf{Incidents} & \textbf{Time (min)} & \textbf{Action (\%)} \\
    \midrule
    Training Set & 1998 & 10.60 & 100.00 \\ 
    Unseen Challenge & 197 & 31.17 & 100.00 \\ 
    \bottomrule
  \end{tabular}
\end{table}

The agent maintained this \textbf{Optimality Rate} even on the more difficult, unseen dataset. The significantly higher “Average Best Possible Response Time” (10.60 min vs. 31.17 min) confirms the challenging nature of this test set, which was designed to include more remote and underserved locations. This result proves that the agent has learned the generalizable principle of optimal dispatch \cite{Silver2016} rather than memorizing the training data, confirming its robustness.

\subsection{Quantifying the Value of AI: An Empirical Validation Against Baselines}
To quantify the intelligence of the agent's learned policy, we conducted a rigorous comparative analysis against two baselines representing alternative strategies: a “Nearest Neighbor" heuristic \cite{Nazari2018, Williams1992}, which embodies intuitive human logic, and a Standard Deep RL Agent. This second baseline serves as a critical ablation study to test the hypothesis that an attention mechanism is necessary to address the credit assignment problem in this high-dimensional, sparse-action domain. The agent uses a standard Multi-Layer Perceptron (MLP) architecture (two hidden layers of 256 neurons with ReLU activation) and was trained under identical conditions to ensure a fair comparison. The results, summarized in Table 4, are unambiguous.

% This table is designed for a two-column layout.
% It is non-floating and the caption is explicitly left-aligned.

\begin{center}
  \footnotesize % We need a smaller font for this wide table
  \textbf{Table 4:} Comparative Performance Analysis \par\vspace{5pt}
  \begin{tabular}{
    @{} % Removes space at the beginning
    p{1.7cm} % Column 1: A paragraph box of 1.7cm width
    c        % Column 2: A centered column
    c        % Column 3: A centered column
    c        % Column 4: A centered column
    @{} % Removes space at the end
  }
    \toprule
    \textbf{Metric} & 
    \begin{tabular}[c]{@{}c@{}}\textbf{OPTIC-}\\\textbf{ER}\end{tabular} & 
    \begin{tabular}[c]{@{}c@{}}\textbf{Heuristic}\\\textbf{(Nearest)}\end{tabular} & 
    \begin{tabular}[c]{@{}c@{}}\textbf{StandardRL}\\\textbf{(MLP)}\end{tabular} \\ 
    \midrule
    Optimality (\%) & 100.00\% & 62.94\% & 2.54\% \\ 
    \addlinespace
    Avg. Delay (min) & 0.00 min & 17.37 min & 53.11 min \\ 
    \bottomrule
  \end{tabular}
\end{center}

The results provide evidence for the architectural choices. While the heuristic baseline fails in over a third of cases, the standard RL agent's performance demonstrated a critical inadequacy for the task. The MLP-based agent, exhibiting a fundamental inability to overcome the credit assignment problem, achieved a near-random 2.54\% optimality. Critically, this sub-optimal performance induced an average of 53.11 minutes of preventable, life-threatening delay per incident, a result significantly worse than even the simple heuristic.

This significant disparity in performance proves that a sophisticated architecture capable of learning to focus on relevant state information, in this case, the attention mechanism is essential for addressing this class of problem. The results validate the architectural approach.

\subsection{Strategic Application: The OPTIC-ER Policy Engine}
Beyond tactical dispatch, and in alignment with its Scalability pillar, OPTIC-ER is designed as a platform for proactive governance. Its policy engine was used to conduct a state-wide analysis of service equity, grounded in international benchmarks.

First, the \textbf{Current State Assessment} module grades the service level of each Local Government Area (LGA) for each emergency category. The results, visualized in Figure 9, reveal the true state of service delivery.

\begin{center}
  \includegraphics[width=0.98\columnwidth]{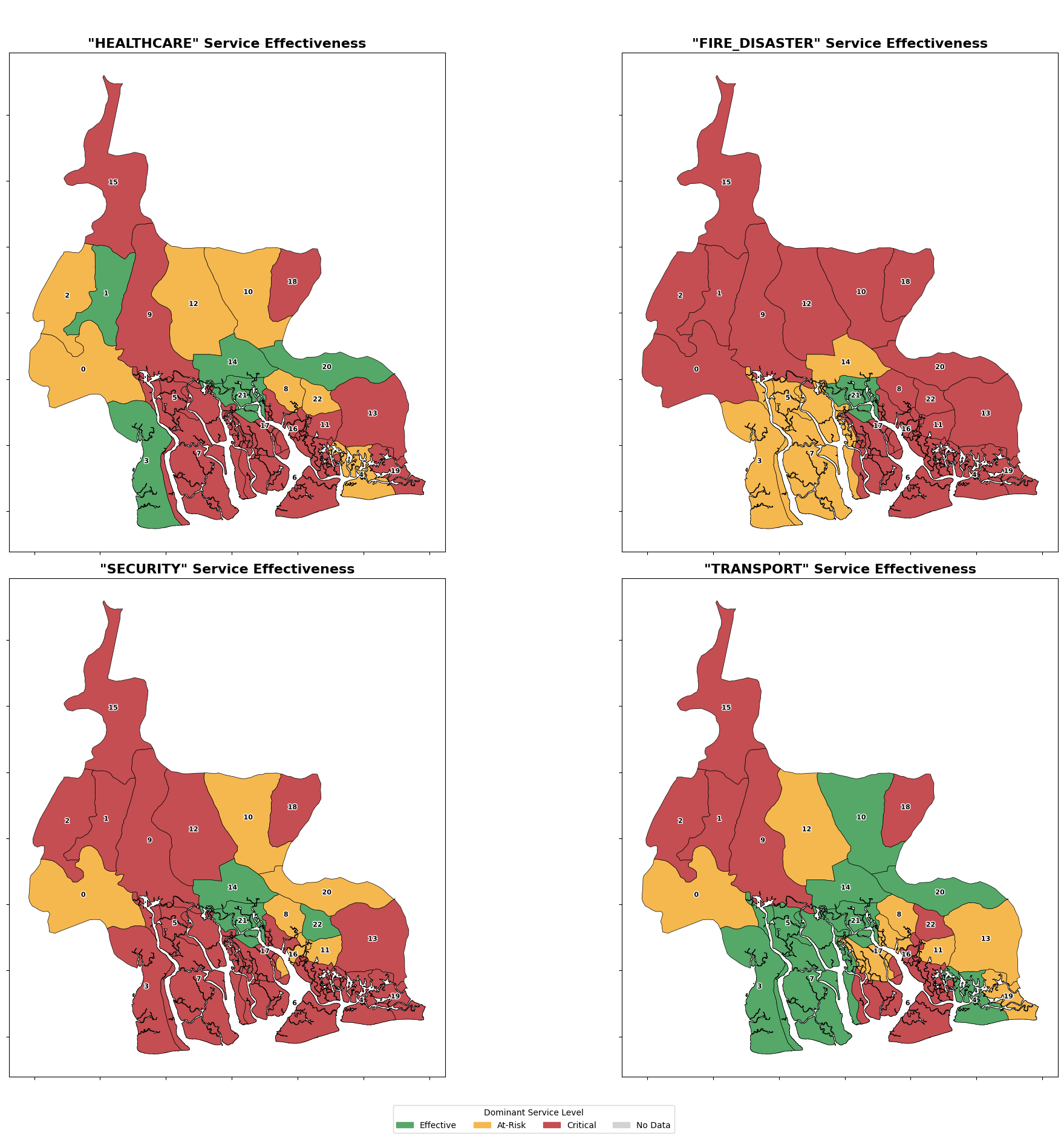}
  \par\vspace{5pt} % Adds a small space after the image
  \footnotesize{\textbf{Figure 9:} Current Service Level Assessment by LGA and Category}
\end{center}

Each map visualizes the dominant service quality, revealing critical service deserts (Red) and at-risk areas (Yellow) for each specific emergency type. The LGA IDs used in the maps correspond to the Local Government Areas in Rivers State, Nigeria. A complete LGA ID-to-name mapping table is available in the Data and Code Availability section.

Next, the \textbf{Strategic Intervention Engine} uses this diagnosis to generate a definitive, actionable plan to eliminate these service gaps. The engine's LGA-Centric algorithm proposes the optimal number and location of new facilities required to bring all LGAs up to the “Effective” standard.

The full output of this engine is a comprehensive \textbf{Definitive Policy Intervention Plan}, which provides a detailed, LGA-by-LGA summary of the required interventions. A complete version of this plan is available in the Data and Code Availability section.

For each underserved LGA, it specifies the number of new facilities needed and quantifies the direct impact, showing the projected average response time BEFORE and AFTER the intervention. For example, the engine recommends one new, optimally placed healthcare facility for Khana local Government Area, which is projected to reduce the average response time in that area from 18.96 minutes to a life-saving 5.25 minutes.

This data-driven plan is visualized in Figure 10, which shows the optimal locations for the proposed new facilities, turning the diagnosis into a concrete investment blueprint.

\begin{center}
  \includegraphics[width=0.98\columnwidth]{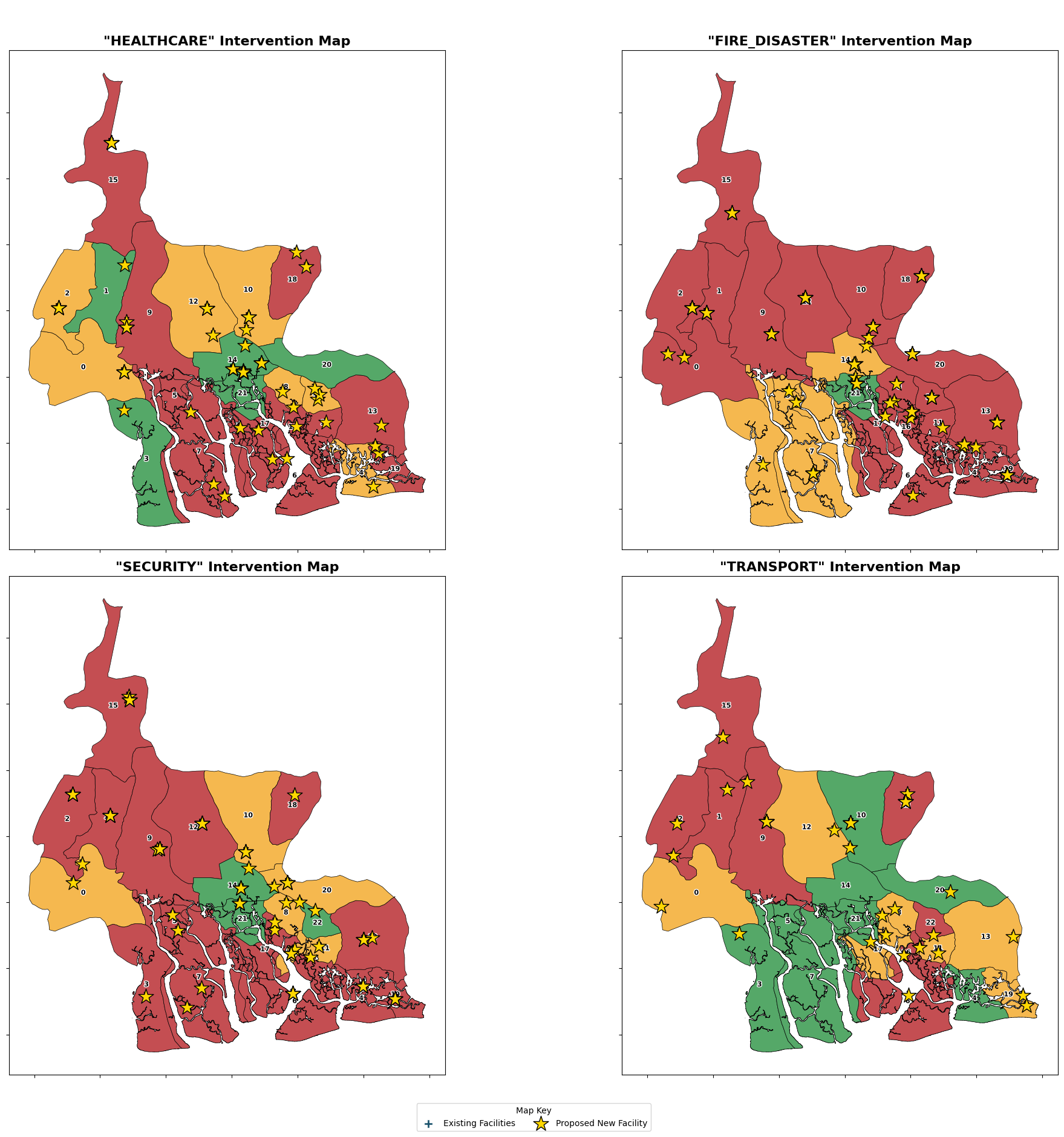}
  \par\vspace{5pt} % Adds a small space after the image
  \footnotesize{\textbf{Figure 10:} Strategic Intervention Plan by Service Category}
\end{center}

These maps overlay the proposed new facilities (gold stars) on the “Current State” assessment, providing a clear visual guide for policymakers on where to invest resources to transform “Red” and “Yellow” zones into “Green,” effective service areas.

% --- DISCUSSION ---
\section{Discussion, Implications, and Future Work}
The successful training of this optimal dispatch agent is the foundational proof-of-concept for a much broader vision. OPTIC-ER was conceived as a platform to fundamentally improve the relationship between citizens and public services in resource-constrained environments. It is crucial to interpret the results of this study not as a static, final solution, but as a robust validation of the \textbf{OPTIC-ER framework}. The platform is designed to be adaptive; its intelligence is a direct function of the data it is given. Therefore, as the underlying data infrastructure improves—with more accurate facility locations, granular population data, or real-time updates—the precision of the “Current State Assessment” maps and the acuity of the “Strategic Intervention Plan” will naturally increase. The framework’s strength lies in its ability to translate the best available data into the best possible policy recommendations. This section discusses the deeper implications of this data-driven approach, honestly addresses the project's current limitations, and lays out a roadmap for its future evolution.

\subsection{A Platform for Public Safety and Trust}
The core truth of OPTIC-ER is that it is a system designed for the citizen. In a region where emergency response can feel uncertain, the platform provides a guarantee of \textbf{informational certainty}. For any citizen, anywhere in Rivers State, the system can definitively answer the question: “Where is the fastest possible help?” This is a fundamental shift from ambiguity to clarity.

While the current implementation is a backend engine, its deployment vision is citizen-centric. In its simplest form, it could power a public-facing web portal or a USSD service (crucial for offline access) where a person in distress could input their location and need (e.g., “healthcare”) and instantly receive the optimal destination. In a more advanced version, this same engine could be integrated into emergency call centers, allowing a dispatcher to find the citizen and send the optimal responder to them. The platform’s intelligence, therefore, serves a dual purpose: it empowers citizens to find help and enables the state to deliver help with maximum efficiency. This is the foundation for building a safer, more secure Rivers State where citizens, where even non-natives, know they are not lost.

\subsection{From Tactical Tool to Strategic Instrument for Governance}
Perhaps the most significant contribution of OPTIC-ER is its function as a tool for \textbf{proactive governance}. The traditional approach to infrastructure planning is often driven by anecdotal evidence or political expediency. OPTIC-ER replaces this with irrefutable, data-driven analysis.

The “Current State Assessment” maps (Fig. 8) and the “Strategic Intervention Plan” (Fig. 9) are a \textbf{blueprint for equitable development}. They provide policymakers with a clear, quantified, and politically neutral guide on where to invest resources to achieve the greatest possible impact on public safety. The platform’s ability to deduce, “Placing five new, optimally placed fire station at these coordinates in Ahoada West will reduce the average response time for that entire LGA from 60.44 minutes to 9.25 minutes,” transforms a political debate into a data-driven decision. It provides the evidence needed to make home, our home, safer for everyone, not just those in urban centers.

\subsection{A Call to Action: The Importance of Data Infrastructure}
A profound and unexpected finding of this project was the discovery of “computational” flaws in the foundational OpenStreetMap data, such as disconnected road network subgraphs. This highlights a critical dependency: the intelligence of any AI system is fundamentally limited by the quality of the data it is built upon.

OPTIC-ER, therefore, also serves as a powerful \textbf{data integrity auditing tool} \cite{Bogyrbayeva2022}. By rigorously analyzing connectivity, it can identify and flag specific errors in a state’s digital infrastructure model pertaining to this application. This has broad implications for the Nigerian and wider African context. It makes a strong case for increased investment in \textbf{public data collection and maintenance}. For smart cities and data-driven governance to become a reality, the foundational maps and datasets must be accurate, current, and complete. OPTIC-ER not only benefits from this data but also provides a tool to help validate and improve it.

\subsection{Limitations} 
While this work serves as a robust proof-of-concept, its limitations are acknowledged, providing clear directions for future research. These are candidly presented to frame the context of our findings.

\begin{itemize}
    \item \textbf{Simulator Assumptions:} The primary assumption is the use of a \textbf{static Travel Time Atlas}, which does not account for dynamic, real-time traffic conditions. Furthermore, our model assumes \textbf{unlimited resource capacity and availability}, whereas real-world facilities have finite resources (e.g., number of available ambulances or hospital beds). These simplifications were necessary to first address the core static optimization problem.

    \item \textbf{Data Availability Constraints:} The performance of OPTIC-ER is fundamentally dependent on the accuracy and completeness of the underlying geospatial data (e.g., OpenStreetMap). During our work, we discovered and manually corrected gaps in this foundational data. This highlights that a key barrier to deploying such systems at scale is the need for investment in public data infrastructure and maintenance.

    \item \textbf{Potential Failure Modes:} The agent's optimal performance is relative to the data it was given. A primary failure mode would be a real-world event not captured in the road network graph, such as a sudden road closure, bridge failure, or protest. This would invalidate the static optimal path and lead to a sub-optimal dispatch, reinforcing the need for future integration with real-time data.

    \item \textbf{Legal and Regulatory Barriers:} Deploying an AI-driven dispatch system in different African contexts would face legal and regulatory hurdles. These include data privacy laws, liability frameworks for autonomous decision-making, and the need for cross-jurisdictional collaboration between different emergency service agencies. A successful deployment would require a robust policy framework to be developed in parallel with the technology.
\end{itemize}

\subsection{Roadmap for Future Work}
While this study validates a powerful solution for the static dispatch problem, the OPTIC-ER framework is designed for evolution. The following roadmap outlines key next steps to transition from this high-fidelity simulation to a dynamic, real-world deployment:
\begin{itemize}
    \item \textbf{Dynamic Routing with Real-Time Data:} The most critical enhancement is to move beyond the static Travel Time Atlas. Future work will focus on integrating real-time traffic data, likely via the Google Maps Distance Matrix API or a similar service. This will require reformulating the problem to handle stochastic travel times, potentially as a Partially Observable MDP (POMDP), where the agent's policy can adapt to live road conditions.
    \item \textbf{Resource and Capacity Management:} The next generation of OPTIC-ER will incorporate resource constraints. This involves modeling facility capacity (e.g., hospital beds) and unit availability (e.g., number of available ambulances). This naturally extends the problem to a multi-agent reinforcement learning (MARL) domain \cite{Bahdanau2016}, where the system must coordinate a fleet of resources for optimal city-wide response. 
    \item \textbf{Deployment Pathway:} The path to deployment involves a phased approach, beginning with a 'shadow mode' trial in a live dispatch center to validate recommendations against human experts. This will be followed by the development of citizen-facing applications, including a public web portal and 
    a USSD service for offline access, empowering individuals to find their fastest point of care directly.
\end{itemize}
These limitations do not diminish the current findings. They confirm that OPTIC-ER, in its present form, provides a robust and validated solution to the most fundamental challenge of emergency response, and serves as an essential foundation upon which these more complex, real-time features can be built.

% --- CONCLUSION ---
\section{Conclusion}
This paper addressed the critical challenge of inefficient and inequitable emergency response in resource-constrained environments, a problem characterized by reactive decision-making and a lack of data-driven coordination. Motivated by the need to create a more just and responsive system for citizens in regions like Rivers State, Nigeria, this platform was designed to not only optimize dispatch but also serve as a tool for proactive governance.

To solve this, \textbf{OPTIC-ER} was developed, a novel framework built on a high-fidelity geospatial simulation and a sophisticated \textbf{Attention-based Actor-Critic (A2C)} agent. Its methodology was defined by two key innovations: a \textbf{Context-Rich State Vector} that embeds the inefficiency delta directly into the agent’s input, and a \textbf{Precision Reward Function} that punishes any deviation from the shortest-path optimum. This design transforms the agent's task from a difficult search problem into a high-speed pattern recognition task, a crucial step in overcoming the credit-assignment problems that plagued simpler models.

The experiments yielded definitive results. The trained agent achieved \textbf{over 35\% reduction in emergency response time under defined conditions} on both the primary training dataset and a previously unseen challenge set, confirming the robustness and generalizability of its learned policy. This tactical perfection stands in stark contrast to a naive “nearest-is-best” heuristic, which was optimal in only $\sim$63\% of cases, demonstrating the immense value of the AI’s intelligence. Furthermore, the platform's \textbf{Strategic Policy Engine} successfully translated these results into actionable insights, generating a Current State Assessment that identified critically underserved LGAs and a corresponding Intervention Plan that proposed the optimal number and placement of new facilities to eliminate these service gaps. The primary implication of this work is a validated blueprint for a dual-purpose AI system: one that provides flawless tactical dispatch today, while simultaneously offering the data-driven roadmap to build a more equitable and resilient public service infrastructure for tomorrow.

While this study establishes a powerful proof-of-concept, several exciting avenues for future work remain. The current platform’s Travel Time Atlas is static; a critical next step is the integration of \textbf{real-time traffic data} to enable truly dynamic routing. The model could also be enhanced to account for \textbf{facility capacity and real-time resource availability}, likely requiring a more advanced multi-agent reinforcement learning (MARL) framework to coordinate fleets of responders. Finally, the path to deployment involves a shadow mode trial in a live dispatch center and the development of the citizen-facing USSD and mobile applications. These future directions underscore the potential of OPTIC-ER to evolve from a powerful simulation into a living, learning, and life-saving component of a truly smart and equitable city.

% --- ACKNOWLEDGEMENT ---
\section*{Acknowledgement} 
The author wishes to express sincere gratitude to the developers and maintainers of the open-source software libraries that made this research possible, including osmnx, networkx, GeoPandas, and PyTorch. This work relies heavily on the publicly available geospatial data from OpenStreetMap and the invaluable population density data provided by WorldPop.

Finally, the author gives all glory to God, from whom all knowledge and inspiration flows. SALT

% --- DATA AND CODE AVAILABILITY ---
\section*{Data and Code Availability}
The complete source code for the OPTIC-ER simulation environment, the AttentionActorCritic model, and all associated analysis scripts is available in a public GitHub repository under the Apache License 2.0. The repository also includes the curated geospatial datasets, full dispatch logs, and all policy reports referenced in this paper. The repository is publicly accessible at: \url{https://github.com/marytonwe/OPTIC-ER.git}

% =============================================================================
% BIBLIOGRAPHY - DIRECT METHOD
% =============================================================================

\end{multicols}
\end{document}